\pdfoutput=1
\documentclass{article}

 \usepackage[nonatbib,preprint]{nips_2018}
\usepackage{graphicx}
\graphicspath{{figure/}}
\usepackage{algorithm}
\usepackage[noend]{algpseudocode}
\usepackage{multirow}      
\usepackage{multicol} 
\usepackage{subfigure}
\usepackage{amsmath}
\usepackage[utf8]{inputenc} 
\usepackage[T1]{fontenc}    

\usepackage{url}            
\usepackage{booktabs}       
\usepackage{amsfonts}       
\usepackage{nicefrac}       
\usepackage{microtype}      
\usepackage{capt-of}
\usepackage{hyperref}       
\title{Hybrid Pruning: Thinner Sparse Networks for Fast Inference on Edge Devices}

\author{
  Xiaofan Xu\thanks{Equal contribution}\qquad Mi Sun Park\footnotemark[1] \qquad Cormac Brick\\
  Movidius, AIPG, Intel\\
  \texttt{xu.xiaofan@intel.com\qquad mi.sun.park@intel.com} \\
}

\begin{document}

\maketitle

\begin{abstract}
We introduce hybrid pruning which combines both coarse-grained channel and fine-grained weight pruning to reduce model size, computation and power demands with no to little loss in accuracy for enabling modern networks deployment on resource-constrained devices, such as always-on security cameras and drones. Additionally, to effectively perform channel pruning, we propose a \textit{fast sensitivity test} that helps us quickly identify the sensitivity of within and across layers of a network to the output accuracy for target multiplier–accumulators (MACs) or accuracy tolerance. Our experiment shows significantly better results on ResNet50 on ImageNet compared to existing work, even with an additional constraint of channels be hardware-friendly number.
\end{abstract}
\vspace{-5mm}
\section{Introduction}
Recently, with the increasing size of large-scale datasets and high-end GPUs, lots of unprecedented large DNN models were developed. Although these modern networks, such as VGG\cite{simonyan2014very}, GoogleNet\cite{szegedy2015going}, DenseNet\cite{huang2017densely} and ResNets\cite{he2016deep} show outstanding classification performance on ImageNet (even better than human), they are not easily deployable on resource-constrained inference devices due to large demands for memory, compute and power. 

Network Pruning is to remove some redundant weights (or channels) which don’t contribute a lot to the output of a network. As a result, it reduces model size and helps preventing over-fitting, and eventually generate sparse (or thinner) model. 
Weight pruning \cite{han2015learning}\cite{guo2016dynamic} show high compression rate on AlexNet (9 to 17.7$\times$) by pruning redundant weight or additionally allowing splicing of previously pruned weights. 
Channel-pruning work \cite{Li2016PruningFF}\cite{Molchanov2016PruningCN}\cite{luo2017thinet}\cite{He_2017_ICCV}\cite{Liu2017learning} naively prune equal percentage of number of channels based on their own calculation of importance of channels. Although the one goal of identifying importance of channels, remove those that are relatively less important and fine-tune the pruned network is the same for all these works, none of these work has considered the fact that each network has different sensitivity within and across layers to output accuracy. Therefore, evenly prune 50\% of channels in all the layers results in significant accuracy drop.  
To overcome the limitation, we propose a fast sensitive test for identifying corresponding number of channels in each layer for a given accuracy tolerance or target MACs. We also apply weight pruning on channel pruned thinner model to further reduce model size and computation. This hybrid pruning eventually generates thinner sparse model. To the best of our knowledge, this is the first to show that the coexistence of multi granularity of sparsity can help significantly reduce resource demands with no significant loss in accuracy, and gives SOTA result on thinner ResNet50 on ImageNet, with channels be multiple of 8. 


\section{Hybrid Pruning}
\label{sec:hybrid}
\begin{figure*}[!htbp]
\vspace{-2mm}
\begin{center}
\includegraphics[width=0.5\linewidth,height=15mm]{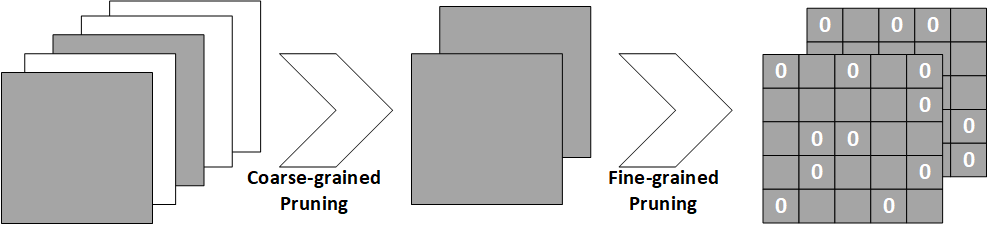}
\end{center}
\caption{Pictorial view of our proposed hybrid pruning}
\label{fig:overview}
\vspace{-2mm}
\end{figure*}

In this work, we applied both coarse-grained channel and fine-grained weight pruning on convolutional layers for various types of neural networks. We call this combination hybrid pruning. Figure \ref{fig:overview} illustrates the process for one convolutional layer using hybrid pruning resulting in a thinner sparse model.

\textbf{Sensitivity-aware channel pruning:} 
In general, one big challenge in channel pruning is that if we naively decide the number of channels to prune away based on target sparsity, it will leads to significant loss in accuracy as shown in  \cite{Li2016PruningFF}\cite{Molchanov2016PruningCN}\cite{luo2017thinet}\cite{He_2017_ICCV}\cite{Liu2017learning} regardless of their filter or channel selection. For ResNet50 on ImageNet, \cite{He_2017_ICCV} and \cite{luo2017thinet} reported that for 50\% compression in parameters gives 1.4 to 1.1 drop @top5 which is significant loss considering this is top5, not even top1. 
As Algorithm \ref{alg:sensitiveytest} shows, our proposed fast sensitivity test helps us quickly identify the sensitivity of each layer in a network for a given accuracy tolerance. The basic idea is to find the upper bound in the percentage of the number of intact output channels in each layer with potential recoverable loss in accuracy with fine-tuning. This was inspired by the observation that when we damage the network too much by pruning more than acceptance, it becomes more difficult to recover its accuracy. From our experiments, accuracy tolerance of 3 to 5\% works well for ResNet50 on ImageNet. 
Furthermore, since we know number of channels and required MACs in an original network and the relative sensitivity of each layer from the test, we can calculate the final number of channels for target MACs by multiplying the pruning percentage with the original channel number, with an additional constraint of channels be hardware-friendly number.
It should be also noted that the sensitivity test itself does not require any training and it only takes 8.86 minutes on 1-CPU Intel Core i7-6850K CPU and 3.38 seconds on GPU GTX-1080-Ti Pascal for conducting this test on ResNet50 on ImageNet. Unlike the existing works AutoML\cite{he2018amc} and\cite{chin2018layer}, our method doesn't require any extra training for new models or meta-learning, and it is compatible with any frameworks. It can help users to quickly identify the pruning channels without large GPU resources.





\vspace{-2mm}
\begin{algorithm}
\caption{Fast sensitivity test for channel pruning}
\scriptsize
\textbf{Input:} Validation data and a dense model $M$.

\textbf{Output:} Pruning percentage for each layer in the model $M$. 
\begin{algorithmic}[1]
\State Threshold accuracy = original dense accuracy - accuracy tolerance (e.g. 3 - 5\%)
\For {each layer in the model $M$}
\State sort output channels based on the sum of absolute weight values 
\For {sparsity percentage in between 30\% - 80\%}
\State channel-wise mask is created based on the current sparsity percentage
\State accuracy = \textbf{Forward} the network with $channel \times mask$
\If{accuracy > Threshold accuracy}
\State continue;
\Else 
\State Record the \textbf{per-layer} percentage and exit the for loop
\EndIf
\EndFor
\State \textbf{end for}
\EndFor
\State \textbf{end for}
\end{algorithmic}
\textbf{Note:} Final number of channels per-layer = round((100\% - pruning \% per-layer) $\times$ original number of channels per-layer), with round(.) to be multiple of 8 or 4. 

\label{alg:sensitiveytest}
\end{algorithm}
\vspace{-2mm}

\textbf{Statistics-aware weight pruning:} 
After generating thinner dense model from sensitivity-aware channel pruning, we apply statistics-aware weight pruning on the thinner model for further reduction in model size and computation. The basic idea is to compute layer-wise weight threshold based on the current statistical distribution of full dense weights in each layer and mask out those weights that are less than the threshold. To do so, we use layer-wise $mask_{l}^{n}$ to represent the binary mask for $l^{th}$ layer at $n^{th}$ iteration, shown in Equation \ref{eq:mask}. This binary mask is dynamically updated in each iteration of training based on the threshold that is computed based on the mean and standard deviation of the weights in each layer with sparsity level controlling factor $\sigma$ ($\sigma$ is the same for all the layers). Since we sparsify weights in forward pass and update full dense weights with the gradients computed with sparse weights during training, it allows previously pruned weights back if it becomes more important ($ |W_{l}^{n}(i,j)| > t_{l}^{n}$), similar to \cite{guo2016dynamic}. Our experiment shows that our statistics-aware method works better than sparsifying all the layers with the same sparsity level.

\begin{equation}
\vspace{-2mm}
mask_{l}^{n}(i,j) = \begin{cases}
0 & \text{ if } \left | W_{l}^{n}(i,j) \right|   <   t_{l}^{n} \\ 
1 & \text{ if } \left | W_{l}^{n}(i,j) \right|  \geq  t_{l}^{n}
\end{cases}
\label{eq:mask}
, \quad  t_{l}^{n} = mean(|W_{l}^{n}|) + std(|W_{l}^{n}|)\times \sigma
\vspace{-2mm}
\end{equation}



\section{Experiments}

In this section, we show the effectiveness of our proposed fast sensitivity test for channel pruning that can give instant speedup on existing hardware, and moreover demonstrate the coexistence of multi granularity of sparsity can help significantly reduce resource demands for fast inference on edge devices. 
For fair comparison, similar to previous work \cite{Li2016PruningFF}\cite{luo2017thinet}, we do not prune the first Conv, the last FC layer and the last layer in each residual block.

\begin{figure}[!htbp]
\vspace{-6mm}
\begin{center}
\includegraphics[width=0.65\linewidth,height=25mm]{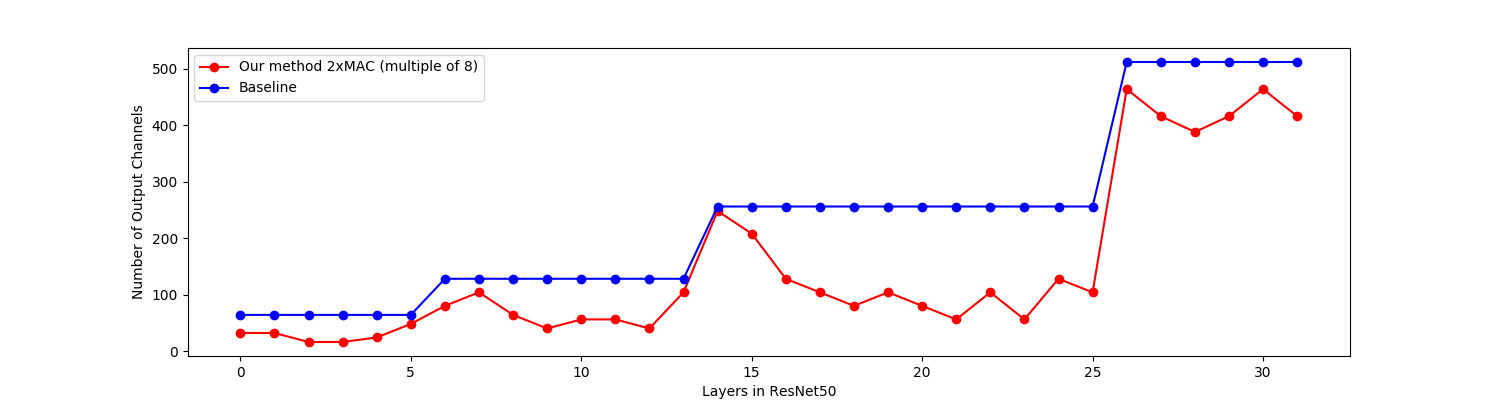}
\end{center}
\vspace{-3.5mm}
\caption{Number of output channels in ResNet50}
\vspace{-2mm}
\label{fig:channelprune50}
\end{figure}

\textbf{ResNet56 on Cifar10:} 
Table\ref{tab:hybridcifar} shows the result of our hybrid pruning on (bottleneck based) ResNet56. 
Our hybrid pruning contains the channel pruning shown in Figure\ref{fig:channelprune56}, it should be noted that in Figure\ref{fig:channelprune56} and \ref{fig:channelprune50}, the first Conv, the last FC and the last layer in each residual block are omitted for clarity, since these layers are intact and have no reduction in output channels, compared to original network (blue dots). The number of survived channels (red dots) in each layer are predetermined from the sensitivity test with accuracy tolerance of 2\% to be conservative in accuracy loss. Furthermore, we added an additional constraint of channels be multiple of 4 for optimal hardware utilization. 
As a result, most of the layers can be pruned up to 75\% with no significant loss in accuracy. It is also observed that those few layers that have increased output channels (compared to previous layer) are more sensitive. 
After achieving 2.4$\times$ instant speedup based on channel pruning, our hybrid pruning gives final 4.5$\times$ reduction in parameters (additional 1.8$\times$ reduction on thinner ResNet56) with less than 1\% accuracy drop. Hybrid pruning is only applied on the convolutional layers to boost the model to 78\% sparsity and it can be further pruned by applying weight pruning on the FC layer.



\begin{table*}[!htbp]
\vspace{-3mm}
	\resizebox{\columnwidth}{!}{%

	\resizebox{0.5\columnwidth}{!}{%
\begin{minipage}[b]{0.5\linewidth}
\centering
\includegraphics[ width=65mm,height=25mm]{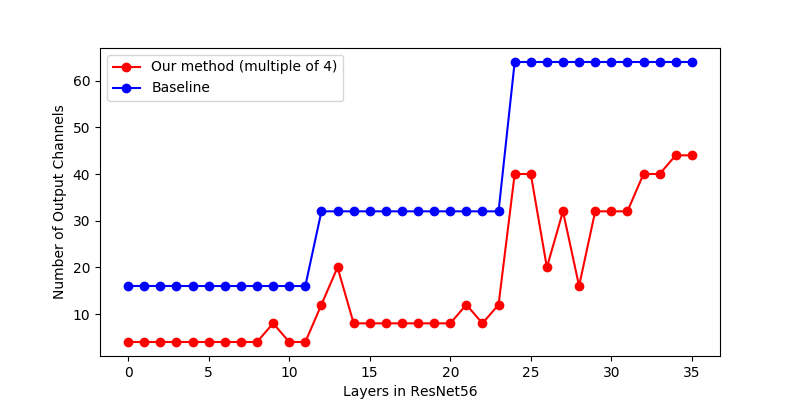}
\captionof{figure}{Number of output channels in ResNet56}
\label{fig:channelprune56}
\end{minipage}\hfill
}
\resizebox{0.5\columnwidth}{!}{%
\begin{minipage}[H]{0.65\linewidth}
\centering
\vspace{-40mm}
\caption{Hybrid pruning on ResNet56 on Cifar10 with comparison to other works}
\scriptsize 
\begin{tabular}{cccc}
\hline
 & \begin{tabular}[c]{@{}c@{}}Top 1 (\%)\\ Baseline $\rightarrow$Pruned\end{tabular} & \begin{tabular}[c]{@{}c@{}}No. of Params\\ Baseline$\rightarrow$Pruned\end{tabular} & \begin{tabular}[c]{@{}c@{}}Sparsity (\%)\\ Baseline$\rightarrow$Pruned\end{tabular} \\ \hline
PF\cite{Li2016PruningFF} & 93.04$\rightarrow$91.31 & 0.73M$\rightarrow$- & 0$\rightarrow$13.7 \\ \hline
SFP\cite{he2018soft} & 93.59$\rightarrow$92.26 & - & - \\ \hline
CP\cite{He_2017_ICCV} & 92.8$\rightarrow$91.8 & - & - \\ \hline
\begin{tabular}[c]{@{}c@{}}Our\\ Channel Pruning\end{tabular} & 93.27$\rightarrow$92.67 & 0.59M$\rightarrow$0.24M & 0$\rightarrow$59 \\ \hline
\begin{tabular}[c]{@{}c@{}}Our\\ Hybrid Pruning\end{tabular} & 93.27$\rightarrow$\textbf{92.48} & 0.59M$\rightarrow$\textbf{0.13M} & 0$\rightarrow$\textbf{78} \\ \hline
\end{tabular}
 
   \label{tab:hybridcifar}
\end{minipage}
}
\vspace{-5mm}
}
\end{table*}

\textbf{ResNet50 on ImageNet:}
Figure\ref{fig:channelprune50} shows the output channels of pruned ResNet50 using ImageNet.The experiment was intended for 2$\times$MAC reduction based on the sensitivity test with a constraint of channels be multiple of 8. 
It is observed that, similar to ResNet56, those few layers that have increased output channels (compared to previous layer) are more sensitive. Interestingly, unlike ResNet56, the last few layers that have the most number of channels are also sensitive. From this observation, we can infer that although network structure looks similar, it might be safe to assume that each network has different sensitivity and it's also different across layers within a network, therefore it's important to have a such technique that can help us to quickly identify the sensitivity of a network. As Table \ref{tab:imagenet} shows, our sensitivity based channel pruning method outperformed the existing naive methods with large margin. Even our hybrid ResNet50 gives better accuracy with less number of parameters (3.7$\times$ reduction in parameters). In addition to parameter storage and bandwidth savings, both key for edge devices, our hybrid model can further boost the performance on any hardware with sparse matrix support.

\begin{table}[!htb]
\vspace{-2mm}
\centering
\caption{Hybrid Pruning on ResNet50 on ImageNet with comparison to other works\label{tab:imagenet}}
\resizebox{\columnwidth}{!}{%
\begin{tabular}{cccccccccc}
\hline
 & \begin{tabular}[c]{@{}c@{}}Top 1 (\%)\\ Baseline\end{tabular} & \begin{tabular}[c]{@{}c@{}}Top 1 (\%)\\ Channel Pruning\end{tabular} & \begin{tabular}[c]{@{}c@{}}Top 1 (\%)\\ Hybrid Pruning\end{tabular} & \begin{tabular}[c]{@{}c@{}}Top 5 (\%)\\ Baseline\end{tabular} & \begin{tabular}[c]{@{}c@{}}Top 5 (\%)\\ Channel Pruning\end{tabular} & \begin{tabular}[c]{@{}c@{}}Top 5 (\%)\\ Hybrid Pruning\end{tabular} & Original No. of Params & \begin{tabular}[c]{@{}c@{}}No. of Params\\ (Coarse-grained Pruning)\end{tabular} & \begin{tabular}[c]{@{}c@{}}No. of Params\\ (Hybrid Pruning)\end{tabular} \\ \hline
CP\cite{He_2017_ICCV} & - & - & - & 92.2 & 90.8 & - & - & 2$\times$ reduction & - \\ \hline
NISP\cite{yu2017nisp} & - & 0.21\% reduction & - & - & - & - & - & 27.12\% reduction & - \\ \hline
SPP\cite{wang2017structured} & - & - & - & 91.2 & 90.4 & - & - & 2$\times$ reduction & - \\ \hline
ThiNet\cite{luo2017thinet} & 72.88 & 71.01 & - & 91.14 & 90.02 & - & 25.5M & 12.38M & - \\ \hline
SFP\cite{he2018soft} & 76.15 & 74.61 & - & 92.87 & 92.06 & - & - & 30\% reduction & - \\ \hline
Ours & 76.01 & \textbf{74.87} & \textbf{74.32} & 92.93 & \textbf{92.43} & \textbf{92.05} & 25.5M & \begin{tabular}[c]{@{}c@{}}17.2M \\ (32.5\% reduction)\end{tabular} & \textbf{\begin{tabular}[c]{@{}c@{}}6.9M \\ (72.9\% reduction)\end{tabular}} \\ \hline
\end{tabular}
}
\vspace{-4mm}
\end{table}

\section{Conclusion}
In this paper, we introduced hybrid pruning that combined both coarse-grained and fine-grained pruning to enable modern networks deployment on edge devices. To the best of our knowledge, this work is the first to show that the coexistence of multi granularity of sparsity can help significantly reduce resource demands with no significant loss in accuracy. 
We also proposed a new \textit{fast sensitivity test} technique, which helps us quickly identify the sensitivity of a network to the output accuracy for a given accuracy tolerance or target MACs.  


\bibliographystyle{ieee}
\bibliography{egbib}

\end{document}